\pdfoutput=1
\documentclass[11pt]{article}

\usepackage{acl}

\usepackage{times}
\usepackage{latexsym}
\usepackage{booktabs}
\usepackage{pifont}    % 用于 \ding
\usepackage{amssymb}   % 用于 \checkmark
\usepackage{amsmath,amssymb,amsfonts}

\usepackage{graphicx}
\usepackage{booktabs}

\usepackage{enumitem}

\usepackage{algorithm}
\usepackage{algorithmic}
\usepackage{fontawesome}  % 放在导言区

\usepackage{microtype}

\usepackage{subcaption}

\usepackage{listings}
\newcommand{\cmark}{\ding{51}}  % 对号
\newcommand{\xmark}{\ding{55}}  % 叉号
\usepackage{amsmath}
\usepackage{mathtools}

\usepackage{booktabs}      % 提供 \toprule, \midrule, \bottomrule 等命令
\usepackage{colortbl}
\definecolor{lightpink}{rgb}{1.0, 0.85, 0.9}

\usepackage{multirow}      % 如果后面要用到 \multirow
\usepackage{caption}       % 可选，用于微调表格标题
\usepackage{siunitx}       % 可选，如果需要对齐数字及百分比

\usepackage[T1]{fontenc}
\usepackage[utf8]{inputenc}

\usepackage{microtype}

\usepackage{inconsolata}

\usepackage{graphicx}

\title{\textsc{THCM-Cal}: Temporal-Hierarchical Causal Modelling with Conformal Calibration for Clinical Risk Prediction}

\author
{
Xin Zhang$^{1}$, Qiyu Wei$^{1}$, Yingjie Zhu$^{2}$, Fanyi Wu$^{1}$, 
Sophia Ananiadou$^{1}$\\
$^{1}$The University of Manchester \quad
$^{2}$Harbin Institute of Technology \quad \\
\texttt{\{xin.zhang-41@postgrad., qiyu.wei@postgrad., fanyi.wu@\}manchester.ac.uk}\\
\texttt{sophia.ananiadou@manchester.ac.uk, zhuyj@stu.hit.edu.cn}
}

\begin{document}
\maketitle
\begin{abstract}
Automated clinical risk prediction from electronic health records (EHRs) demands modeling both structured diagnostic codes and unstructured narrative notes. However, most prior approaches either handle these modalities separately or rely on simplistic fusion strategies that ignore the directional, hierarchical causal interactions by which narrative observations precipitate diagnoses and propagate risk across admissions. In this paper, we propose \textsc{THCM-Cal}, a Temporal-Hierarchical Causal Model with Conformal Calibration. Our framework constructs a multimodal causal graph where nodes represent clinical entities from two modalities: Textual propositions extracted from notes and ICD codes mapped to textual descriptions. Through hierarchical causal discovery, \textsc{THCM-Cal} infers three clinically grounded interactions: intra-slice same-modality sequencing, intra-slice cross-modality triggers, and inter-slice risk propagation. To enhance prediction reliability, we extend conformal prediction to multi-label ICD coding, calibrating per-code confidence intervals under complex co-occurrences. Experimental results on MIMIC-III and MIMIC-IV demonstrate the superiority of \textsc{THCM-Cal}.
%尽管以往有对多模态信息的融合，但是他们不仅过少关注模态之间的交互，也没有对文本信息充分利用
% Experimental results on five
% real-world datasets demonstrate the superiority
% of the proposed method
% (+4.7\% AUROC, +6.2\% F1), with ablation studies confirming that cross-modality triggers contribute over 15\% F1 gain. 
\end{abstract}

% \section{Introduction}
\section{Introduction}

Accurate clinical risk prediction from Electronic Health Records (EHRs)~\cite{evans2016electronic} is essential for enabling timely clinical interventions and improving treatment effects \cite{Choi2016RETAIN,miotto2016deep}. EHRs comprise two complementary data modalities: Structured diagnostic codes drawn from the International Classification of Diseases (ICD) \cite{Choi2017GRAM, bodenreider2004unified} and Unstructured narrative notes that chronicle patient observations and interventions over time \cite{Huang2019ClinicalBERT}. Leveraging both modalities can enhance automated code assignment~\cite{sun2024enhanced}, risk stratification~\cite{tsai2025harnessing}, and downstream decision support \cite{Lu2021CGL,Xu2023VecoCare}.

\begin{figure}[!t]
\centering
\includegraphics[width=0.49\textwidth]{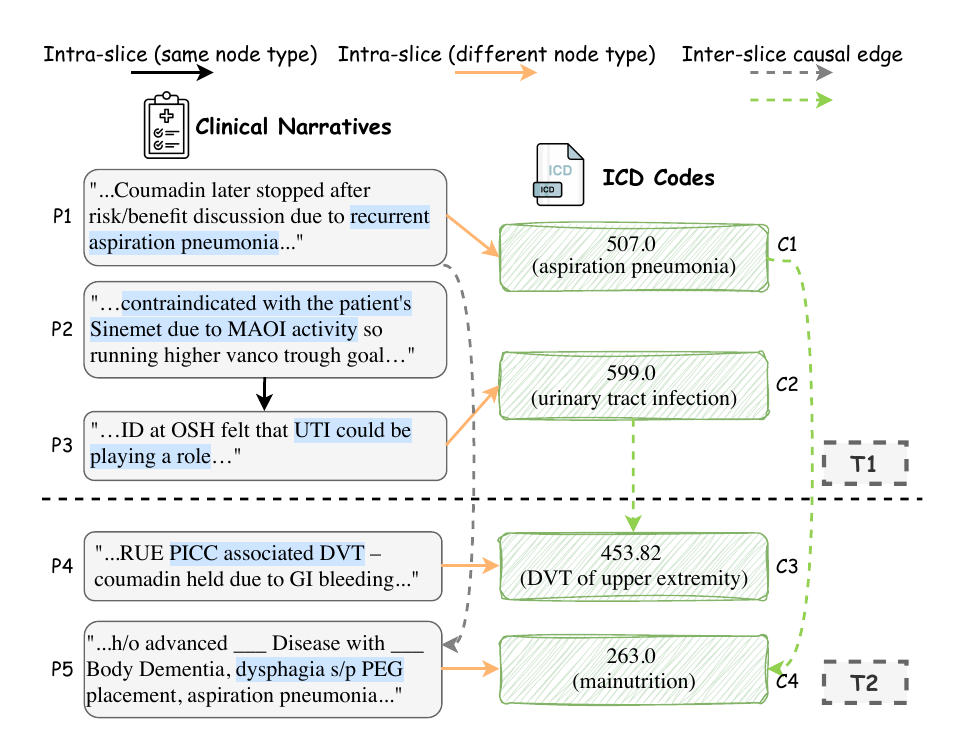}
% \caption{Temporal causal graph of key textual observations (blue) and ICD codes across hospitalizations T1 and T2. Edge types: intra-slice sequencing (P2→P3), intra-slice cross-modal triggers (P1→C1), and inter-slice propagation (C1→C4).}
\caption{An illustrative two‐slice causal graph over textual propositions and ICD nodes.
Three directed edge types are learned: 
intra‐slice sequencing (P$_2 ~ \to ~ $P$_3$), 
intra‐slice cross‐modality triggers (P$_1 ~ \to ~$C$_1$), and 
inter‐slice propagation (C$_1 ~ \to ~ $C$_4$).}
\label{fig:intro}
\end{figure}

\begin{table*}[!t]
  \centering
  % \caption{Comparison of THCM-CAL against representative methods.}
  \caption{Overview of representative clinical risk prediction methods, comparing their input modalities, support for causal structure discovery, uncertainty estimation capabilities, and temporal hierarchical modeling.}

  \label{tab:method_comparison}
  \setlength{\tabcolsep}{4pt}
  \resizebox{0.98\linewidth}{!}{
    \begin{tabular}{@{} llcccc @{}}
      \toprule
      Method & Modalities & Causal Discovery & Uncertainty & Temporal Hi \\
      \midrule
      % RETAIN \cite{Choi2016RETAIN}           & Codes         & \xmark                 & \xmark            & \xmark    \\
      CAML \cite{Mullenbach2018CAML}         & Text          & \xmark                 & \xmark            & \xmark    \\
      ZAGNN \cite{Zheng2019ZAGNN}            & Text          & \xmark                 & \xmark            & \xmark    \\
      DistilBioBERT \cite{Sharan2022DistilBioBERT} & Text    & \xmark                 & \xmark            & \xmark    \\
      BioMedLM \cite{Li2023BioMedLM}         & Text          & \xmark                 & \xmark            & \xmark    \\
      GatorTron \cite{Wei2021GatorTron}      & Text          & \xmark                 & \xmark            & \xmark    \\
      DKEC \cite{Chen2023DKEC}               & Code+Text     & \xmark                 & \xmark            & \xmark    \\
      Chet~\cite{lu2022context}              & Code+Text     & Dynamic graph          & \xmark            & \xmark    \\
      CDANs \cite{Ferdous2023CDANs}          & Codes         & Instantaneous \& lagged edges & \xmark      & \cmark    \\
      CACHE \cite{Xu2022CACHE}               & Codes         & Hypergraph-based       & \xmark            & \xmark    \\
      COMPOSER \cite{Gottesman2021COMPOSER}  & Codes         & \xmark                 & Single-task CP    & \xmark    \\
      \midrule
      \textbf{THCM-CAL (Ours)}               & \textbf{Code+Text} & \textbf{Intra/inter-slice, cross-modality} & \textbf{Multi-label CP} & \cmark \\
      \bottomrule
    \end{tabular}
  }
\end{table*}

% Automated diagnosis‐code assignment from EHRs is typically formulated as a multi‐label text classification task, characterized by an exponential label space and long‐tail distributions across ICD codes. 
% Early approaches addressed one modality at a time: label‐wise attention over clinical narratives such as CAML \cite{Mullenbach2018CAML} and ZAGNN \cite{Zheng2019ZAGNN} highlighted salient text spans but ignored structured code context, while code‐centric models like RETAIN \cite{Choi2016RETAIN} and GRAM \cite{Choi2017GRAM}) applied attention to historical code sequences yet overlooked narrative semantics. The advent of large pre‐trained transformers such as Chet \cite{lu2022context}, DistilBioBERT \cite{Sharan2022DistilBioBERT}, BioMedLM \cite{Li2023BioMedLM} and GatorTron \cite{Wei2021GatorTron} markedly improved textual representation learning, but these models remain task‐agnostic during pre‐training and often neglect inter‐code dependencies and narrative‐to‐code triggers. Domain‐knowledge frameworks like DKEC \cite{Chen2023DKEC} incorporate external ontologies to inform predictions but rely on static graph structures that cannot adapt to patient‐specific trajectories.
% However, existing methods treat text and codes as co‐occurring signals rather than dynamically interacting processes. Table~\ref{tab:method_comparison} provides an overview of representative methods.

Previous approaches as shown in Table~\ref{tab:method_comparison} fall into two broad categories. On one hand, text-centric models such as CAML~\cite{Mullenbach2018CAML} and ZAGNN~\cite{Zheng2019ZAGNN} leverage label-wise attention over narrative notes but entirely ignore structured code context, while code-focused methods like RETAIN~\cite{Choi2016RETAIN} and GRAM~\cite{Choi2017GRAM} attend only to historical ICD sequences and overlook the rich semantics of free-text observations. On the other hand, recent transformer-based and text-driven frameworks including Chet~\cite{lu2022context}, DistilBioBERT~\cite{Sharan2022DistilBioBERT}, BioMedLM~\cite{Li2023BioMedLM}, GatorTron~\cite{Wei2021GatorTron} and DKEC~\cite{Chen2023DKEC} improve representation power or inject external knowledge yet still treat text and codes as static co-occurrences. They remain task-agnostic during pretraining, neglect fine-grained narrative-to-code triggers and rely on fixed graph structures that cannot adapt to patient-specific time-varying causal relationships.

Taking the patient trajectory in Figure~\ref{fig:intro} as an example. In the first admission (T$_1$), the note reports a medication contraindication (P$_2$) that prompts a reassessment of the urinary tract infection etiology (P$_3$), capturing an intra‐slice sequencing dependency. The description “recurrent aspiration pneumonia” (P$_1$) directly precipitates the assignment of ICD code J69.0 (C$_1$), exemplifying an intra‐slice cross‐modality trigger. Finally, the diagnosis C$_1$ in T$_1$ increases the likelihood of a related complication (C$_4$) in the subsequent admission (T$_2$), demonstrating inter‐slice risk propagation. These patterns highlight three critical dimensions of clinical causality that current models overlook: how narrative observations temporally trigger diagnoses, how events in one hospitalization propagate risk to the next, and how causal dependencies span hierarchical temporal scales. Existing approaches either operate on a single slice or treat text–code interactions as undirected associations, and thus fail to capture these directed, modality‐aware causal mechanisms.  

% However, all the aforehead approaches treat modalities as static rather than dynamically interacting, thereby overlooking three critical dimensions of clinical causality: 
% i) how textual observations temporally trigger specific diagnoses;
% ii) how intra-hospitalization events propagate risks to subsequent hospitalizations; and
% iii) how causal dependencies operate at hierarchical temporal levels.
% Figure~\ref{fig:intro} exemplifies these challenges through a patient’s hospitalization trajectory. Multimodal interactions manifest as three distinct patterns: i) Intra-slice same-modality dependencies (e.g., P2→P3 in T1, where a medication contraindication leads to reassessment of UTI etiology), ii) Intra-slice cross-modality triggers (e.g., P1→C1 in T1, where "recurrent aspiration pneumonia" justifies ICD code J69.0), and iii) Inter-slice propagation (e.g., C1→C4 across T1–T2). Existing methods either focus on single time slices or model cross-modal interactions as undirected correlations, failing to capture these directed, modality-aware causal mechanisms.

% 信息抽取、节点表示、图构建 & 融合、校准

To address these gap, we propose \textsc{THCM-Cal}, a Temporal–Hierarchical Causal Model with Conformal Calibration for Clinical Risk Prediction. Our framework proceeds in four stages:
First, we segment each admission’s narrative into diagnostically relevant sections;
Second, we encode both propositions and code descriptions with BERT and project them into a shared embedding space to form textual and code nodes.
Third, we sample intra-slice same-modality, intra-slice cross-modality, and inter-slice propagation edges using Gumbel–Softmax~\cite{jang2016categorical} with acyclicity constraints; then fuse these edges via graph~\cite{rossi2020temporal} message passing to produce per-admission embeddings.
Finally, we apply split conformal prediction~\cite{shafer2008tutorial} to the multi-label probabilities, generating valid confidence sets at a desired coverage level even under complex code co-occurrences.
% Experiments on MIMIC-III and eICU demonstrate that \textsc{THCM-Cal} outperforms state-of-the-art baselines by +5.5\% Precision@10 and +5.22\% Recall@10. 
Our key contributions are:
\begin{itemize}
  \item We propose a causal framework to unify intra‐visit sequencing, cross‐modality triggers, and cross‐visit propagation in a hierarchical graph.
  \item We develop a split‐conformal calibration method that provides distribution‐free uncertainty guarantees on the prediction of diagnostic codes.
  \item We demonstrate that explicit causal modeling of multimodal interactions yields obvious gains in performance, interpretability, and robustness, with ablations showing each module contributes.
\end{itemize}

\begin{figure*}[!t]
\centering
\includegraphics[width=1\textwidth]{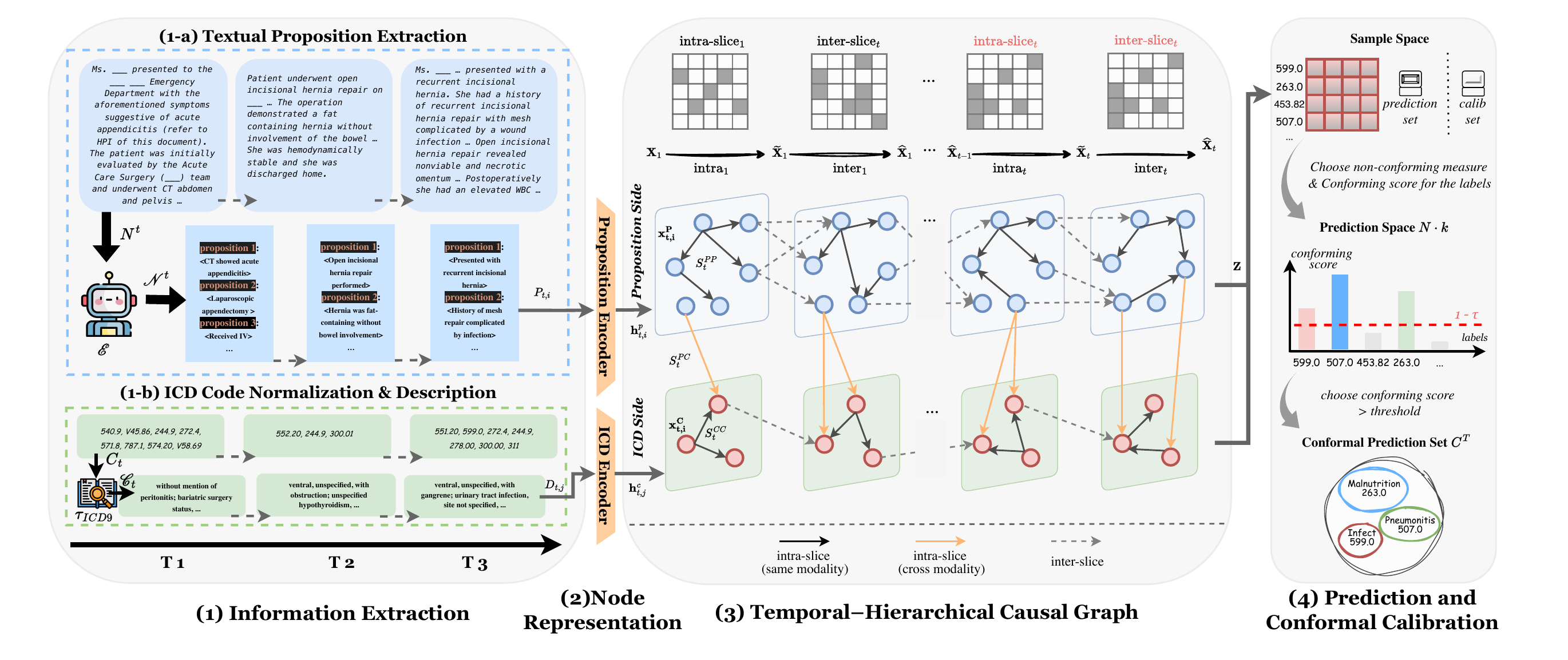}
% \caption{Overview of \textsc{THCM-Cal}, a four-stage pipeline for clinical risk prediction and uncertainty quantification. 
% (1) Information extraction: segment each admission’s notes into diagnostic sections, extract atomic propositions via GPT-3.5, and map raw ICD codes to their textual descriptions. 
% (2) Node representation: encode propositions and code descriptions with BERT and project into a shared embedding space. 
% (3) Temporal–hierarchical causal graph: sample intra-slice same-modality (P$\to$P, C$\to$C), intra-slice cross-modality (P$\to$C), and inter-slice propagation edges using Gumbel-Softmax with acyclicity constraints, then fuse these graphs via message passing to produce per-admission embeddings. 
% (4) Conformal calibration: apply split conformal prediction to the predicted multi-label probabilities to generate valid confidence sets at a user‐specified coverage level.}
\caption{Overview of \textsc{THCM-Cal}. A four-stage pipeline for clinical risk prediction, which consists of: (1) Extracting diagnostic propositions and normalize ICD descriptions, (2) Embeding nodes with BERT, (3) Building and fuse a temporal–hierarchical causal graph via Gumbel–Softmax and message passing, and (4) Appling split conformal prediction for calibrated multi‐label ICD coding.}
% \caption{Overview of \textsc{THCM-Cal}, a four-stage pipeline for clinical risk prediction, which consists of 
% (1) Information Extraction,
% (2) Node Representation,
% (3) Temporal–Hierarchical Causal Graph Construction and 
% (4) Conformal Calibration.}

\label{fig:Framework}
\end{figure*}

\section{Related Work}
We categorize prior work into three areas: text-centric diagnosis prediction, temporal causal discovery, and uncertainty quantification. 
% Table~\ref{tab:method_comparison} provides an overview of representative methods, highlighting their modalities, support for causal discovery, uncertainty estimation, and temporal modeling.

\subsection{Text-Centric Diagnosis Prediction}
Automated diagnosis-code prediction from free-text clinical narratives is commonly formulated as a multi-label text classification task mapping notes to sets of ICD codes. Early label-wise attention networks such as CAML \cite{Mullenbach2018CAML} and ZAGNN \cite{Zheng2019ZAGNN} assign each code its own attention mechanism to highlight relevant text spans and capture label co-occurrence and hierarchy. More recent transformer-based approaches including DistilBioBERT \cite{Sharan2022DistilBioBERT}, BioMedLM \cite{Li2023BioMedLM} and GatorTron \cite{Wei2021GatorTron} learn rich contextual embeddings via large-scale pre-training but remain task-agnostic and often disregard explicit code dependencies at fine-tuning. Domain-knowledge frameworks such as DKEC \cite{Chen2023DKEC} integrate external ontologies to inform prediction but rely on static graph structures that may not reflect patient-specific interactions between text and codes. Despite their strengths, these methods overlook directional triggers between narrative findings and subsequent codes.

\subsection{Causal Discovery in Medical}
Causal discovery in EHRs has mainly focused on structured codes. SemDBN \cite{Wang2018SemDBN} employs ontology-augmented Bayesian networks for sepsis onset prediction yet excludes unstructured narratives. FGES-based techniques \cite{Shen2020CausalEHR} improve edge orientation in static cohorts, and CDANs \cite{Ferdous2023CDANs} extend causal discovery to time series by modeling lagged dependencies. CACHE \cite{Xu2022CACHE} applies hypergraph learning and counterfactual inference to structured codes but does not capture how textual propositions precipitate specific diagnoses. As a result, these methods cannot reveal cross-modal, directional relationships between narrative observations and ICD assignments.

% However, these methods cannot capture cross-modality triggers (e.g., how “recurrent aspiration” in text justifies ICD code J69.0). 

\subsection{Uncertainty Quantification in Clinical Prediction}
Conformal prediction provides finite-sample coverage guarantees but has been applied mainly to single-label clinical forecasting. COMPOSER \cite{Gottesman2021COMPOSER} and NeuroSep CP \cite{Zhou2025NeuroSep} generate confidence intervals for sepsis onset and temporal risk trajectories, respectively, under an assumption of label independence. Sepsyn OLCP \cite{Zhou2025Sepsyn} adapts conformal methods to online monitoring but does not address multi-label ICD coding, where code co-occurrences induce complex uncertainty dependencies. Consequently, existing frameworks do not yield valid per-code confidence sets for multi-label diagnosis prediction.

% \textbf{Summary of Gaps.} As Table~\ref{tab:method_comparison} illustrates, existing methods exhibit three limitations: (1) multimodal models treat code-text interactions as static correlations rather than directional triggers, (2) causal discovery ignores narrative-driven diagnoses, and (3) uncertainty quantification lacks multi-label calibration. Our work addresses these through hierarchical causal graphs and conformal calibration.

\section{\textsc{THCM-Cal}}
\label{sec:method}

\subsection{Task Definition}
\label{sec:task}

We consider a cohort of $S$ patients, each with up to $T$ chronological hospital admissions
\(
A^1, A^2, \dots, A^T.
\)
For the $t$-th admission, we denote the raw data by
\(
A^t = \bigl(\tau^t,\;N^t,\;C^t\bigr),
\)
where $\tau^t\in\mathbb{R}$ is the admission timestamp, $N^t$ is the free-text clinical note, and $C^t$ is the set of recorded ICD-9 codes.
After Stage 1 (“Information Extraction”), each admission is transformed into
\(
\mathcal A^t = \bigl(\tau^t,\;\mathcal C^t,\;\mathcal N^t,\;\mathcal P^t,\;\mathcal D^t\bigr),
\)
where $\mathcal N^t = \mathcal E (N^t)$ is the cleaned clinical narrative, $\mathcal C^t = \tau_{\mathrm{ICD9}}(C^t)$ is the normalized set of ICD-9 codes, $\mathcal P^t$ is the extracted propositions and $\mathcal D^t$ is the labels.
Given the history of the first $T-1$ admissions,
\(
\{\mathcal A^1,\dots,\mathcal A^{T-1}\},
\)
our objective is to predict the ICD-9 code set at the next admission:
\(
\hat{\mathcal C}^T \;=\; f_{\mathrm{ICD}}\bigl(\mathcal A^1,\dots,\mathcal A^{T-1}\bigr).
\)
We train and evaluate $f_{\mathrm{ICD}}$ under the standard multi-label framework, comparing $\hat{\mathcal C}^T$ to the ground-truth $\mathcal C^T$.

% \subsection{Task Definition}
% Assume there are \(S\) patients and each has \(T\) admissions. We consider each patient’s hospital trajectory as a sequence of \(T\) admissions
% \(
% A^1, A^2, \dots, A^T.
% \)
% For the \(t\)-th admission we define:
% \(
% A^t = \bigl(\tau^t,\;\mathcal{C}^t,\;\mathcal{N}^t\bigr),
% \)
% where \(\tau^t\) is the admission timestamp, \(\mathcal{C}^t\) is the set of ICD-9 codes from raw codes \({C}^t\), \(\mathcal{N}^t\) is the cleaned clinical narrative extracted from the raw note \(N^t\).
% Given the first \(T-1\) admissions \(\{A^1,\dots,A^{T-1}\}\), we aim to predict for the final admission \(A^T\):
% \begin{align*}
%   &\hat{\mathcal{C}}^T = f_{\mathrm{ICD}}\bigl(A^1,\dots,A^{T-1}\bigr), \\
%   &\hat y_{\mathrm{HF}} = f_{\mathrm{HF}}\bigl(A^1,\dots,A^{T-1}\bigr),
% \end{align*}
% where the ground-truth heart-failure label is
% \(
% y_{\mathrm{HF}}
% = \mathbb{I}\bigl\{\mathcal{C}^T \cap \mathcal{D}_{\mathrm{HF}}\neq\emptyset\bigr\},
% \)
% and \(\mathcal{D}_{\mathrm{HF}}\) is the predefined set of heart-failure ICD-9 codes.  

\subsection{Overview of \textsc{THCM-Cal}}

% Figure~\ref{fig:Framework} illustrates \textsc{THCM-Cal}’s end-to-end pipeline. Given raw EHR records, the first stage performs information extraction: clinical notes are segmented into atomic propositions using large language model, and ICD codes are mapped to human-readable descriptions via an ICD-9 lookup. In the second stage, each proposition and code description is encoded by BERT, producing contextual embeddings that serve as graph node features. The third stage constructs a temporal-hierarchical causal graph: intra-admission edges capture proposition sequencing and code co-occurrence, cross-modal edges link observations to diagnoses, and inter-admission edges propagate information across time. A graph fusion network then performs message passing over these edges and aggregates node representations into a compact embedding for each admission. Finally, the conformal calibration stage applies split conformal prediction to the predicted ICD code probabilities, yielding confidence sets that ensure the user’s specified coverage. Each stage builds directly on the previous outputs to deliver accurate, interpretable, and reliable clinical risk predictions.

Figure~\ref{fig:Framework} summarizes the four connected modules that compose \textsc{THCM-Cal}. Starting from raw EHRs, we first segment each admission’s free‐text note into diagnostically meaningful sections and invoke a large language model to extract a set of atomic propositions. In parallel, all recorded ICD codes are normalized to ICD-9 and replaced with their canonical textual descriptions. 
In the second module, every proposition and code description is fed through BERT and projected into a common embedding space, yielding a rich set of node feature vectors. These vectors form the inputs to our third, core component: we assemble a temporal-hierarchical causal graph by sampling three kinds of directed relationships—within‐admission links among propositions and among codes, cross‐modal links from propositions to codes, and across‐admission propagation edges. Edge selection is performed via Gumbel-Softmax under acyclicity constraints, and the resulting multi‐slice graph is combined through a graph fusion that propagates messages and pools node representations into a compact embedding per admission.
Finally, the fourth stage applies split conformal prediction to the model’s multi‐label outputs, producing calibrated confidence sets that respect a user‐specified coverage level. By chaining these modules—information extraction, node encoding, causal graph construction with fusion, and conformal calibration—\textsc{THCM-Cal} delivers interpretable risk estimates with rigorous uncertainty quantification.

\subsection{Stage 1: Information Extraction}

We start from the raw clinical note \(N^t\) and the raw ICD set \({C}^t\), and extract the cleaned narrative \(\mathcal{N}^t\) together with normalized ICD-9 codes \(\mathcal{C}^t\) via $\mathcal N^t = \mathcal E (N^t)$ and $\mathcal C^t = \tau_{\mathrm{ICD9}}(C^t)$.

\paragraph{Text segmentation \& scoring}  
Let \(\mathcal{H}=\{h_1,\dots,h_{|\mathcal H|}\}\) be the set of recognized section headings (e.g., “Chief Complaint”, “History of Present Illness”). We split \(N^t\) into sections
\(
\mathcal{S}=\{\,s_i=\mathrm{SEGMENT}(N^t,h_i,h_{i+1})\mid h_i,h_{i+1}\in\mathcal{H}\}.
\)
If \(\mathcal{S}=\emptyset\), let \(\mathcal{S}=\{N^t_{[1:4000]}\}\).  
Let \(\mathrm{count}_{\texttt{mg}}(s)\) denote the number of occurrences of the substring “mg” in \(s\), \(\mathcal{K}\) denotes a predefined set of medical keywords (e.g., “fever”, “cough”, “pain”) that are strongly indicative of diseases or diagnoses. For each segment \(s_i\in\mathcal S\), we extract a feature vector
\(\mathbf f(s_i) = [f_1(s_i),\,f_2(s_i),\,f_3(s_i),\,f_4(s_i)]^\top\), where
\begin{align*}
f_1(s_i)&= \bigl|\mathcal K\cap s_i\bigr|,\\
f_2(s_i)&=\mathbb I\{\,\texttt{``Diagnosis:''}\in \mathrm{Text}(s_i)\},\\
f_3(s_i)&=\min\bigl(\mathrm{count}_{\texttt{mg}}(s_i),\,9\bigr),\\
f_4(s_i)&=\mathbb I\{\mathrm{SentCount}(s_i)>2\}.
\end{align*}
We define the linear scoring function
\(
\mathrm{score}(s_i)\;=\;\mathbf w^\top \mathbf f(s_i),~
\mathbf w=(2,\,5,\,1,\,3)^\top,
\)
and then sort \(\mathcal S\) by \(\mathrm{score}(\cdot)\) in descending order.  The top-\(K\) segments are retained:
\(
\mathcal S^* = \bigl\{\,s_i\in\mathcal S\mid \mathrm{rank}_{\mathcal S}\bigl(\mathrm{score}(s_i)\bigr)\le K\bigr\}
\),where $K=3$.
\paragraph{Atomic propositions extraction}  
Let \(\mathcal E\colon s\mapsto\mathcal P_s\) denote the GPT-3.5–based extractor that maps a text segment \(s\) to a set of “atomic propositions.”  We then define
\[
\mathcal P^t 
= \mathrm{uniq}\Bigl(\bigcup_{s\in\mathcal S^*}\mathcal P_s\Bigr)
= \{\,P_{t,1},\,P_{t,2},\,\dots,\,P_{t,n_t}\},
\]
where \(\mathrm{uniq}(\cdot)\) removes duplicates and enforces an fixed ordering.  These \(P_{t,i}\) serve as the proposition-node set for Stage 2.
\paragraph{ICD code normalization}  
We normalize the raw ICD codes \(C^t\) via
\(
\mathcal C^t = \tau_{\mathrm{icd9}}\bigl(C^t\bigr),
\)
and retrieve their human-readable labels by
\(
\mathcal D^t = \bigl\{\delta(c)\mid c\in\mathcal C^t\bigr\} = \{D_{t,1},\dots,D_{t,J_t}\},
\)
where \(\tau_{\mathrm{icd9}}\) is a standard ICD-9 mapping, \(\delta\) is the lookup from code to description and \(D_{t,j}\) is the ICD‐description strings.  
Consequently, the complete output of Stage 1 for admission \(t\) is
\[
\mathcal A^t 
= \bigl(\tau^t,\;\mathcal C^t,\;\mathcal N^t,\;\mathcal P^t,\;\mathcal D^t\bigr).
\]

\subsection{Stage 2: Node Representation}
\label{sec:stage2}

\paragraph{BERT Encodings.}  
Let 
\(
\mathrm{Enc}_{\mathrm{BERT}}: \mathcal T \to \mathbb R^d
\)
denote the BERT\citep{DBLP:conf/naacl/DevlinCLT19} mapping from any text token sequence to its \([CLS]\) embedding of dimension \(d\).  We write
\(
\mathbf h^P_{t,i} = \mathrm{Enc}_{\mathrm{BERT}}\bigl(P_{t,i}\bigr),\quad
\mathbf h^C_{t,j} = \mathrm{Enc}_{\mathrm{BERT}}\bigl(D_{t,j}\bigr).
\)
Thus \(\mathbf h^P_{t,i},\,\mathbf h^C_{t,j}\in\mathbb R^d\).

\paragraph{Modality‐Specific Projections.}  
To bring propositions and code descriptions into a shared \(d'\)-dimensional space, we apply two learnable linear projections with nonlinearity:
\begin{align*}
\mathbf x^P_{t,i}
&= \mathrm{Proj}_P(\mathbf h^P_{t,i})
= \phi\bigl(W_P\,\mathbf h^P_{t,i} + b_P\bigr),
\\
\mathbf x^C_{t,j}
&= \mathrm{Proj}_C(\mathbf h^C_{t,j})
= \phi\bigl(W_C\,\mathbf h^C_{t,j} + b_c\bigr),
\end{align*}
where
\(
W_P,\;W_C \in \mathbb R^{d'\times d}, 
\quad
b_P,\;b_C \in \mathbb R^{d'},
\)
and \(\phi(\cdot)\) denotes an element‐wise activation (e.g.\ ReLU) optionally combined with dropout.

\paragraph{Feature Matrix Assembly.}  
Finally, we concatenate all proposition‐ and description‐level vectors into the admission‐level feature matrix
\[
\mathbf X^t
=\bigl[\mathbf x^P_{t,1},\dots,\mathbf x^P_{t,I_t},\,
  \mathbf x^C_{t,1},\dots,\mathbf x^C_{t,J_t}\bigr]
\;\in\;\mathbb R^{d'\times ND_t},
\] where \(ND_t=I_t+J_t\). This \(\mathbf X^t\) serves as the node‐feature input to Stage 3.

\subsection{Stage 3: Temporal–Hierarchical Causal Graph}

% In this stage, we learn three types of within‐admission causal structure and across‐admission temporal dependencies, then propagate information via graph convolutions to obtain per‐admission embeddings.
Inspired by the temporal causal graph construction in RealTCD \cite{li2024realtcd} framework 
, we further extend their approach by building three types of within-admission causal graphs and modeling across-admission temporal dependencies, then propagate information via graph convolutions to obtain per-admission embeddings.

\paragraph{Constructing causal adjacency matrices.}  
To capture causal and temporal dependencies, we construct a slice-wise causal graph at each admission \(t\). Define the Gumbel‐Softmax sampling operator
\(
\mathrm{GS}\bigl(E;\tau\bigr)
=\mathrm{softmax}\bigl((E+G)/\tau\bigr), G_{ij}\sim\mathrm{Gumbel}(0,1).
\)
For each admission \(t\), we parameterize three intra‐slice logit matrices \(\mathbf E_t^{PP},\mathbf E_t^{PC},\mathbf E_t^{CC}\in\mathbb R^{ND_t\times ND_t}\) (with \(ND_t=I_t+J_t\), represents the total number of nodes at time $t$) and obtain adjacency blocks by
\(
\mathbf S_t^{PP}=\mathrm{GS}\bigl(\mathbf E_t^{PP};\tau\bigr),
\mathbf S_t^{PC}=\mathrm{GS}\bigl(\mathbf E_t^{PC};\tau\bigr),
\mathbf S_t^{CC}=\mathrm{GS}\bigl(\mathbf E_t^{CC};\tau\bigr).
\)
We then form the block‐structured intra‐slice adjacency
\(
\mathbf A_t
=\begin{pmatrix}
\mathbf S_t^{PP} & \mathbf S_t^{PC}\\
\mathbf 0            & \mathbf S_t^{CC}
\end{pmatrix}.
\)

To ensure that each intra‐slice adjacency \(\mathbf A_t\) defines a directed acyclic graph, we adopt the continuous acyclicity penalty from NOTEARS~\cite{zheng2018dags}:
\(
h(\mathbf A_t)
=\operatorname{tr}\bigl(\exp(\mathbf A_t \circ \mathbf A_t)\bigr)\;-\;ND_t,
\)
where \(\mathbf A_t \circ \mathbf A_t\) is the Hadamard square of \(\mathbf A_t\) and \(ND_t=I_t+J_t\) is the total number of nodes.  We add \(\lambda_{\mathrm{acyc}}\;h(\mathbf A_t)\) to the overall loss, which vanishes if and only if \(\mathbf A_t\) is acyclic.
In parallel, to promote sparse and clinically interpretable graphs, we include an \(\ell_1\) penalty on the sampled adjacency blocks:
\(
\lambda_{\ell_1}\,\sum_{x,y\in\{P,C\}}\bigl\lVert \mathbf S_t^{}\bigr\rVert_1.
\)
Similarly, inter‐slice logits \(\mathbf E_t^{\mathrm{inter}}\) produce
\(
\mathbf S_t^{\mathrm{inter}}
=\mathrm{GS}\bigl(\mathbf E_t^{\mathrm{inter}};\tau\bigr),
\)
which models influences from slice \(t\) to slice \(t+1\).

\paragraph{Graph Fusion and Temporal Message Passing.}  
Let \(\mathbf X_t\in\mathbb R^{d'\times ND_t }\) be the matrix whose rows are \(\{\mathbf x^P_{t,i}\}_{i=1}^{I_t}\) followed by \(\{\mathbf x^C_{t,j}\}_{j=1}^{J_t}\).  Define two graph‐convolution operators:
\begin{align*}
&\mathrm{GC}_{\mathrm{intra}}(\mathbf{S},\mathbf{X};\mathbf{W})=\mathrm{ReLU}(\mathbf{S}\,\mathbf{X}\,\mathbf{W}) + \mathbf{X},\\
&\mathrm{GC}_{\mathrm{inter}}(\mathbf{S},\mathbf{X};\mathbf{W})=\mathbf{S}\,\mathbf{X}\,\mathbf{W} + \mathbf{X}.
\end{align*}
We perform:
\begin{align*}
\widetilde{\mathbf X}_t
&=\mathrm{GC}_{\mathrm{intra}}\bigl(\mathbf A_t,\;\mathbf X_t;\;\mathbf{W}^{(1)}\bigr),\\
\widehat{\mathbf X}_{t+1}
&=\mathrm{GC}_{\mathrm{inter}}\bigl(\mathbf S_t^{\mathrm{inter}},\;\widetilde{\mathbf X}_t;\;\mathbf{W}^{(2)}\bigr).
\end{align*}
Finally, we pool the updated intra‐slice features and project:
\(
\mathbf Z_t
=\mathrm{MLP}\Bigl(\bigl[\mathrm{Mean}(\widetilde{\mathbf X}_t^{P});\,\mathrm{Mean}(\widetilde{\mathbf X}_t^{C})\bigr]\Bigr),
\)
where \(\mathrm{Mean}(\cdot)\) averages row‐vectors and \(\widetilde{\mathbf X}_t^{P},\widetilde{\mathbf X}_t^{C}\) denote the proposition‐ and code‐node partitions of \(\widetilde{\mathbf X}_t\).  Concatenating \(\{\mathbf Z_t\}_{t=1}^T\) yields the trajectory embedding \(\mathbf Z\in\mathbb R^{T\times k}\), which feeds into the downstream prediction and calibration modules.  

\subsection{Stage 4: Prediction and Conformal Calibration}

\paragraph{Prediction}  
Let \(\mathbf Z\in\mathbb R^{T\times k}\) be the trajectory embedding, where \(T\) is the number of admissions and \(k\) is the embedding dimension.  Define  
\(
\mathbf W_o\in\mathbb R^{k\times L},\quad
\mathbf b_o\in\mathbb R^L,,
\) where \(\mathbf 1\in\mathbb R^T\ )\)is all‐ones vector.
where \(L\) is the number of target labels.  We compute the logit matrix
\(
\mathbf Y
= \mathbf Z\,\mathbf W_o \;+\;\mathbf 1\,\mathbf b_o^\top
\;\in\;\mathbb R^{T\times L},
\)
and apply the element‐wise sigmoid
\(
\sigma(x)=\frac{1}{1+e^{-x}}
\quad\Longrightarrow\quad
\hat{\mathbf P}=\sigma(\mathbf Y)\;\in\;(0,1)^{T\times L},
\)
with \(\hat P_{t,j}=\sigma(Y_{t,j})\).

% \paragraph{Training loss}  
% We train by minimizing the binary cross‐entropy over all admissions \(t\) and labels \(j\):
% \begin{align*}
% \mathcal L_{\mathrm{BCE}}
% = -\sum_{t=1}^T \sum_{j=1}^L &\Bigl[y_{t,j}\,\log\hat P_{t,j}\\
% &+ \bigl(1-y_{t,j}\bigr)\,\log\bigl(1-\hat P_{t,j}\bigr)\Bigr].
% \end{align*}
% The full objective also includes the Stage 3 acyclicity penalty and sparsity regularization:
% \begin{align*}
% \mathcal L
% = &\mathcal L_{\mathrm{BCE}}
% + \lambda_{\mathrm{acyc}}\sum_{t=1}^T h\bigl(\mathbf A_t\bigr)\\
% &+ \lambda_{\ell_1}\sum_{t=1}^T \sum_{x,y\in\{p,c\}}
% \bigl\lVert\mathbf S_t^{xy}\bigr\rVert_1,
% \end{align*}
% where \(h(\mathbf A_t)\) is the NOTEARS acyclicity term introduced in Stage 3.

\paragraph{Training loss}  
We train by minimizing the focal loss over all admissions \(t\) and labels \(j\):
\begin{align*}
\mathcal L_{\mathrm{FL}}
= &-\sum_{t=1}^T \sum_{j=1}^L \Bigl[\alpha\,y_{t,j}\,(1-\hat P_{t,j})^\gamma\log\hat P_{t,j}\\
&+(1-\alpha)\,(1-y_{t,j})\,\hat P_{t,j}^\gamma\log\bigl(1-\hat P_{t,j}\bigr)\Bigr],
\end{align*}
where \(\alpha\) and \(\gamma\) are the focal‐loss balancing and focusing parameters.  
The full objective also includes the Stage 3 acyclicity penalty and sparsity regularization:
\begin{align*}
\mathcal L
= &\mathcal L_{\mathrm{FL}}
+ \lambda_{\mathrm{acyc}}\sum_{t=1}^T h\bigl(\mathbf A_t\bigr)\\
&+ \lambda_{\ell_1}\sum_{t=1}^T \sum_{x,y\in\{p,c\}}
\bigl\lVert\mathbf S_t^{xy}\bigr\rVert_1,
\end{align*}
where \(h(\mathbf A_t)\) is the NOTEARS acyclicity term introduced in Stage 3.

\paragraph{Conformal Calibration}  
We partition the data into a proper training set and a calibration set \(\mathcal D_{\mathrm{calib}}\).  For each calibration pair \((t,j)\in\mathcal D_{\mathrm{calib}}\), define the nonconformity score
\(
\alpha_{t,j} = 1 - \hat P_{t,j}.
\)
Let \(\{\alpha_{(1)}\le\cdots\le\alpha_{(N)}\}\) be these \(N=|\mathcal D_{\mathrm{calib}}|\) scores in ascending order, and set
\(
\tau = \alpha_{\bigl\lceil (N+1)(1-\epsilon)\bigr\rceil}.
\)
At test time for admission \(t\), the conformal prediction set is
\[
\hat C_t = \bigl\{\,j : 1 - \hat P_{t,j} \le \tau\bigr\},
\]
which guarantees the marginal coverage  
\(
\Pr_{(t,j)\sim\text{test}}\bigl(y_{t,j}\in\hat C_t\bigr)\ge 1-\epsilon.
\)

% \subsection{Implementation Details}
\section{Experiments}

To rigorously evaluate \textsc{THCM-Cal}, we conduct comprehensive experiments across several baseline clinical language models and benchmark its diagnostic‐prediction performance against state‐of‐the‐art methods.

\begin{table*}[!t]
  \centering
  \small
  \label{tab:results}
  \resizebox{1\linewidth}{!}{
  \begin{tabular}{@{}lccccc|ccccc@{}}
    \toprule
    \multirow{2}{*}{\textbf{Model}} 
      & \multicolumn{5}{>{\columncolor{gray!15}}c|}{\textbf{MIMIC-III}}
      % \rowcolor{gray!15} \multicolumn{4}{c}{\textbf{\textit{Dataset: MIMIC-IV}}}
      & \multicolumn{5}{>{\columncolor{gray!15}}c|}{\textbf{MIMIC-IV}} \\
    \cmidrule(lr){2-6} \cmidrule(l){7-11}
      & \textbf{AUROC}  & \textbf{Precision@10} & \textbf{Recall@10} & \textbf{Precision@20} & \textbf{Recall@20}
      & \textbf{AUROC}  & \textbf{Precision@10} & \textbf{Recall@10} & \textbf{Precision@20} & \textbf{Recall@20} \\
    \midrule
    CAML%~\cite{Mullenbach2018CAML}                   
      & 93.43 & 20.61 & 18.15 & 14.98 & 25.74 
      & 95.49 & 20.61 & 27.38 & 14.03 & 35.52 \\
    ZAGCNN%~\cite{Zheng2019ZAGNN}                      
      & 89.88 & 17.50 & 15.51 & 12.61 & 21.63 
      & 93.96 & 21.10 & 28.13 & 14.33 & 36.17 \\
    GatorTron%~\cite{Wei2021GatorTron}                
      & \textbf{95.02*} & 20.53 & 17.72 & 15.03 
      & 25.66 
      % & 94.59 & 22.87 & 27.29 & 15.47 & 35.46 \\
      & 95.90 & 22.87 & 30.19 & 15.47 & 38.81 \\
    DistilBioBERT%~\cite{Sharan2022DistilBioBERT}     
      & 94.99 & 20.18 & 17.85 & 14.98 & 25.62 
      & \textbf{95.99*} & 22.59 & 29.83 & 15.33 & 38.50 \\    
    Chet%~\cite{lu2022context}                        
      & 60.17 & 24.52 & 18.82 & 18.61 & 27.71 
      & 60.79 & 19.06 & 26.65 & 12.62 & 34.11 \\
    DKEC%~\cite{Chen2023DKEC}                         
      & 94.47 & 20.64 & 17.92 & 14.55 & 24.47 
      & 94.34 & 20.03 & 26.50 & 13.61 & 34.59 \\
    BioMedLM%~\cite{Li2023BioMedLM}                   
      & 93.66 & 20.56 & 17.84 & 14.90 & 25.38 
      & 95.14 & 20.72 & 27.68 & 14.09 & 35.32 \\
    \midrule
    \rowcolor{lightpink}
    \textsc{THCM-Cal}% -- BERT              
      % & \textbf{*} & \textbf{*} & \textbf{*} & \textbf{*} & \textbf{*} 
      & 92.07 & \textbf{30.02*} & \textbf{24.04*} & \textbf{21.47*} & \textbf{33.16*} 
      & 94.91 & \textbf{28.83*} & \textbf{37.03*} & \textbf{18.66*} & \textbf{46.04*}
      \\
    -- w/o BERT          
      & 86.54 & 20.26 & 17.97 & 14.34 & 25.07
      & 89.24 & 20.24 & 25.44 & 13.98 & 33.66  \\
    -- w/o ICD         
      & 91.16 & 23.17 & 17.58 & 17.38 & 25.82 
      & 92.61 & 20.99 & 27.67 & 14.39 & 36.58 \\
    -- w/o Proposition 
      & 91.97 & 26.07 & 20.55 & 19.27 & 29.25  
      & 93.87 & 22.43 & 28.99 & 15.44 & 38.49 \\
    -- w/o ConCalib         
      & 91.84 & 26.03 & 20.68 & 19.29 & 29.61 
      & 94.33 & 25.36 & 32.63 & 16.94 & 42.03
      % & 94.49 & 24.87 & 32.39 & 16.53 & 41.61
      \\
    \bottomrule
  \end{tabular}
  }
  % \caption{Comparison of multi‐label ICD code prediction performance for representative baselines and \textsc{THCM-Cal} variants on MIMIC‐III and MIMIC‐IV. Metrics reported are AUROC, Recall@10, Recall@20, Precision@10, and Precision@20.}
\caption{Comparison of multi‐label ICD code prediction performance for representative baselines and \textsc{THCM-Cal} variants on MIMIC‐III and MIMIC‐IV. Metrics reported are AUROC, Precision@10, Recall@10, Precision@20, and Recall@20. Cells highlighted in pink denote our \textsc{THCM‐Cal} method, and boldface values marked with ‘‘*’’ indicate the best performance across all methods, ``w/o'' means without.}

  \label{tab:main_exp}
\end{table*}

\subsection{Experimental Setup}

\paragraph{Dataset} 
We evaluate \textsc{THCM-Cal} on two standard EHR benchmarks. For MIMIC-III~\cite{johnson2016mimic}, we include all patients with at least two hospitalizations (7192 patients, 11980 admissions), and for MIMIC-IV~\cite{johnson2023mimic}, due to the risk of label leakage from discharge summaries in the original MIMIC-IV notes, we merge the original MIMIC-IV dataset with MIMIC-IV-Ext-BHC\footnote{\url{https://physionet.org/content/labelled-notes-hospital-course/1.1.0/}}, which is a meticulously cleaned and standardized corpus of clinical notes (Labeled Clinical Notes Dataset for Hospital Course Summarization), and apply the same two hospitalization filter (17526 patients, 38346 admissions). We explicitly omit discharge summaries in the ICD task to prevent label leakage. 
Our objective is multi label ICD-9 prediction, where the first \(N-1\) visits are used to predict the full set of ICD-9 codes at visit \(N\). Each cohort is split by patient into 70\% train, 10\% validation, and 20\% test sets. In the Conformal Calibration stage we use the validation set for calibration.

% \paragraph{Dataset} We evaluate \textsc{THCM-Cal} on two publicly available EHR benchmarks. On MIMIC-III~\cite{johnson2016mimic}, we include all patients with at least two admissions (7 500 subjects, 20 000 visits), and on MIMIC-IV~\cite{johnson2023mimic} we apply the same filtering (85 000 subjects, 250 000 visits), omitting discharge summaries for the ICD task to avoid label leakage. We address two tasks: (1) multi‐label ICD-9 prediction, in which admissions \(1\ldots N-1\) are used to predict the full set of ICD-9 codes at admission \(N\); and (2) heart‐failure detection, a binary classification of whether any I50× code appears at \(N\). Each dataset is split by patient into 70 \% training, 10 \% validation, and 20 \% testing, we use validation set caformal set.

% \paragraph{Baselines} We compare \textsc{THCM-Cal} against structured‐code models RETAIN~\cite{Choi2016RETAIN}, T-LSTM~\cite{Baytas2017TLSTM}, and HiTANet~\cite{Luo2020HiTANet}; selected multimodal approaches Chet~\cite{Lu2022Chet} and CGL~\cite{Lu2021CGL}; the clinical‐reasoning framework CARER~\cite{Nguyen2024CARER}; the domain‐knowledge method DKEC~\cite{Ge2024DKEC}; and the causal learning baseline CACHE~\cite{Xu2022CACHE}. 

\subsection{Baselines}
We compare \textsc{THCM-Cal} against the following state‐of‐the‐art models for multi‐label ICD prediction:
\textbf{GatorTron}~\cite{Wei2021GatorTron}: a 345M‐parameter transformer pretrained on large‐scale EHR narratives to capture clinical language nuances.  
\textbf{DistilBioBERT}~\cite{Sharan2022DistilBioBERT}: a compact 66M‐parameter BERT distilled on biomedical text, offering a lightweight yet effective encoder for clinical notes.  
\textbf{BioMedLM}~\cite{Li2023BioMedLM}: a 2.7B‐parameter language model trained on diverse biomedical corpora, providing rich domain knowledge for downstream classification.
\textbf{CAML}~\cite{Mullenbach2018CAML}: employs per‐label convolutional filters and attention to highlight text spans most relevant to each ICD code. 
\textbf{ZAGCNN}~\cite{Zheng2019ZAGNN}: integrates the ICD code hierarchy via graph convolutions to enable zero‐ and few‐shot code prediction.  
\textbf{Chet}~\cite{lu2022context}: models multi‐label diagnosis prediction as a sequence‐to‐set generation task using a transformer augmented with clinical event encodings.  
\textbf{DKEC}~\cite{Chen2023DKEC}: incorporates external medical ontologies and domain rules into a multi‐label classifier to enforce code consistency and improve rare code recall.

\paragraph{Parameter Setup} 
Our methods use BERT for medical text embedding with hidden dimensions $d = 768$ for both proposition and ICD code representations. Our hierarchical temporal causal model employs a 2-layer architecture with residual connections and layer normalization. We train using Adam optimizer (learning rate $1 \times 10^{-4}$, weight decay $1 \times 10^{-5}$) with batch size 16 and dropout 0.1. Early stopping is applied with patience of 5 epochs over a maximum of 50 epochs. Each admission is represented with a maximum of 50 propositions and 30 ICD codes. The causal structure employs Gumbel-Softmax temperature annealing from 1.0 to 0.1 to enforce DAG constraints. Implementation uses PyTorch on NVIDIA A100 GPUs.

\subsection{Main Results}
Table~\ref{tab:main_exp} reports AUROC, Precision@10, Recall@10, Precision@20 and Recall@20 for the multi‐label ICD‐9 prediction task on MIMIC‐III and MIMIC‐IV. We highlight three key findings:

(1) On MIMIC‐III, \textsc{THCM‐Cal} achieves 30.02\% Precision@10 and 24.04\% Recall@10, representing gains of 5.50 and 5.22 percentage points over the strongest baseline (Chet: 24.52\% / 18.82\%). On MIMIC‐IV, \textsc{THCM‐Cal} attains 28.83\% Precision@10 and 37.03\% Recall@10, improvements of 6.24 and 7.20 points over the best baseline (DistilBioBERT: 22.59\% / 29.83\%). These results underscore the benefit of our narrative‐to‐code attention coupled with hierarchical causal modeling, which elevates the most critical diagnoses into the top‐K predictions even under distribution shift.

(2) \textsc{THCM‐Cal} delivers 33.16\% Recall@20 on MIMIC‐III (versus 25.74\% for CAML and 24.47\% for DKEC), and 46.04\% Recall@20 on MIMIC‐IV (versus 35.52\% for CAML and 34.59\% for DKEC), yielding improvements of 7.42 and 10.52 points, respectively. This boost is driven by our dynamic causal graph component, which explicitly captures cross‐modal and temporal triggers, ensuring that less frequent but clinically important comorbidities are correctly retrieved.

(3) The gap in AUROC between \textsc{THCM‐Cal} and other methods is relatively small, with \textsc{THCM‐Cal} maintaining 92.07\% on MIMIC‐III and 94.91\% on MIMIC‐IV. This stability reflects the effect of our conformal calibration in adjusting confidence estimates for domain shifts, together with the causal structure that guards against spurious correlations.

Together, these results confirm that combining rich contextual embeddings, fine‐grained proposition extraction, and explicit causal modeling can produce more effective code rankings and broader coverage of clinically relevant diagnoses.

\begin{figure*}[ht]
  \centering
  \begin{subfigure}[b]{0.492\textwidth}
    \centering
    \includegraphics[width=\textwidth]{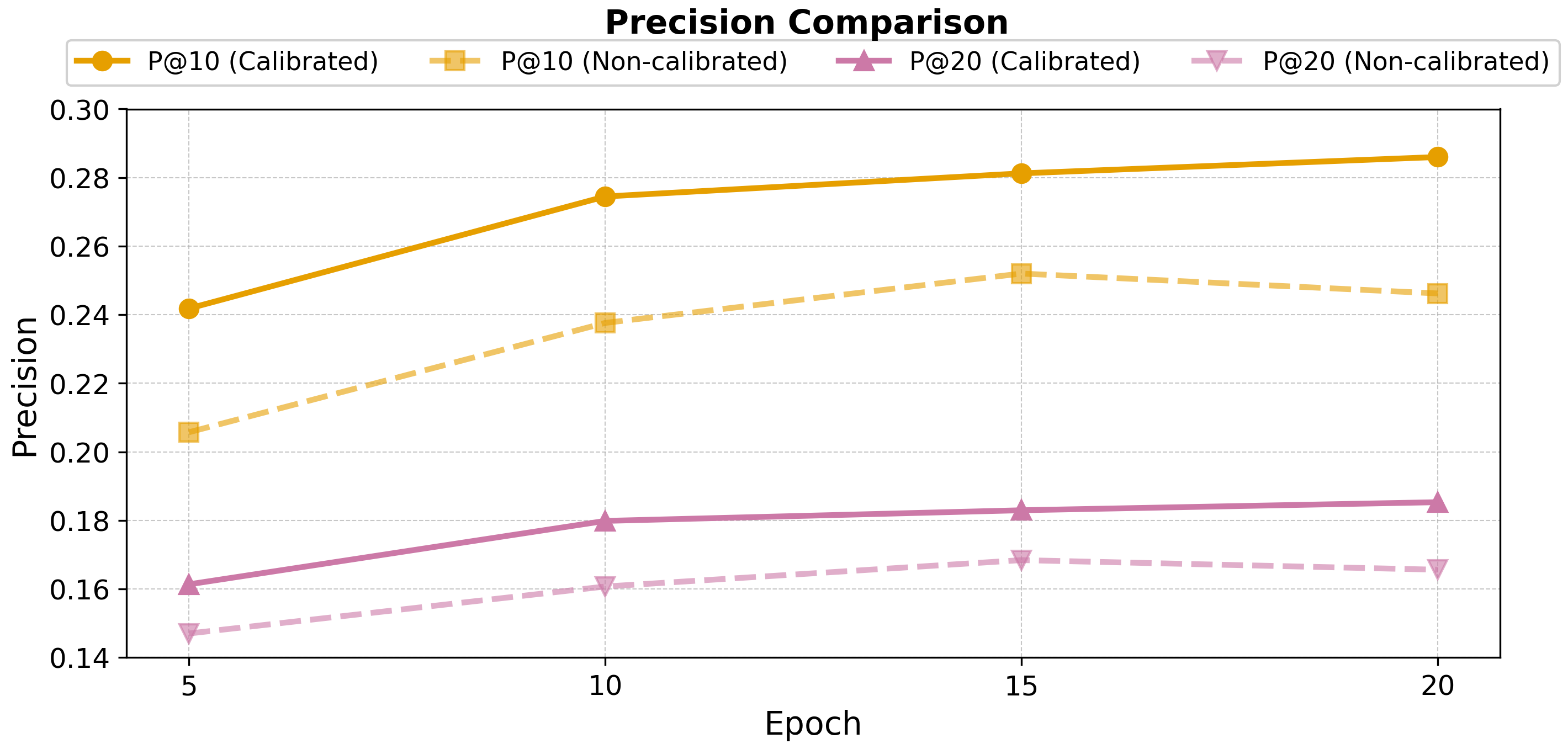}
    % \caption{} 
    % \label{fig:AB-Ca1}
  \end{subfigure}
  \hfill
  \begin{subfigure}[b]{0.492\textwidth}
    \centering
    \includegraphics[width=\textwidth]{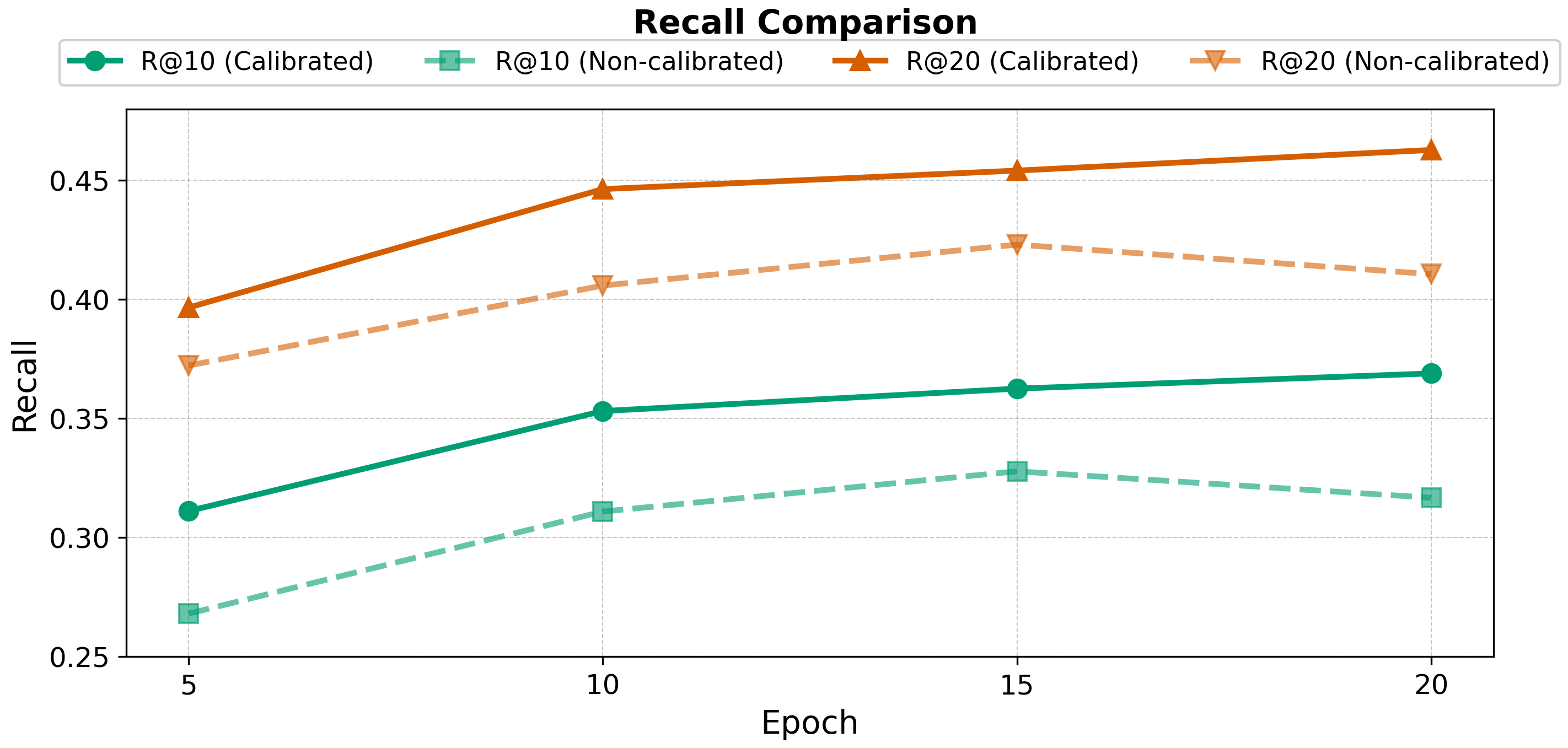}
    % \caption{}
    % \label{fig:AB-Ca2}
  \end{subfigure}
  % \caption{111}
  \caption{Comparison of metrics with and without split conformal calibration on MIMIC-IV.}
  % \caption{Comparison of top-K precision and recall with and without split conformal calibration across training epochs on MIMIC-IV. Calibrated curves (solid) consistently outperform non-calibrated ones (dashed).}
  \label{fig:calibration_effect}
  \label{fig:AB-Ca-combined}
\end{figure*}

% \subsection{Ablation Study}
% To quantify the contribution of each component, we perform an ablation study (Table~\ref{tab:main_exp}) by removing: (i) BERT encoding in favor of simple indexing (“– w/o BERT”); (ii) ICD code–side causal modeling (“– w/o ICD”); (iii) proposition‐side causal modeling (“– w/o Proposition”); and (iv) conformal calibration (“– w/o ConCalib”). Removing any of these components causes performance to decline across all metrics. In particular, replacing contextual embeddings with indexing (– w/o BERT) yields the largest drop, with MIMIC III Precision@20 falling from 20.58\% to 14.34\%, MIMIC IV from 16.94\% to 13.98\% , which demonstrates that rich, contextual text representations are essential for distinguishing fine‐grained clinical details. Eliminating the ICD code‐side causal graph (– w/o ICD) reduces Recall@20 by over 2 points on MIMIC‐III (31.33\% to 29.15\%) and by more than 3 points on MIMIC‐IV (42.03\% to 38.70\%), indicating that explicit modeling of inter‐code dependencies is key to recovering less frequent comorbidities. Finally, skipping conformal calibration (– w/o ConCalib) results in a modest AUROC decrease to 91.00\%, confirming that score recalibration is necessary to maintain stable discrimination under dataset shift.  

\begin{table}[ht]
\centering
\resizebox{0.85\columnwidth}{!}{%
\begin{tabular}{lccc}
\toprule
\rowcolor{gray!15} \multicolumn{4}{c}{\textbf{\textit{Dataset: MIMIC-III}}} \\
\cmidrule(lr){1-4}
Configuration      & Cov.\,$\uparrow$   & MIW\,$\downarrow$   & IE\,$\uparrow$      \\
\midrule
\textsc{THCM-Cal}           & 0.9006   & 0.1087  & 9.1990   \\
-- w/o Proposition             & 0.8980   & 0.1137  & 8.7979   \\
% noprob             & 0.8979   & 0.1164  & 8.5918   \\
% optimized          & 0.8979   & 0.1162  & 8.6062   \\
-- w/o ICD            & 0.8937   & 0.1144  & 8.7412   \\
\midrule
\rowcolor{gray!15} \multicolumn{4}{c}{\textbf{\textit{Dataset: MIMIC-IV}}} \\
\cmidrule(lr){1-4}
Configuration      & Cov.\,$\uparrow$   & MIW\,$\downarrow$   & IE\,$\uparrow$      \\
\midrule
% noprob2            & 0.9005   & 0.0898  & 11.1341  \\
\textsc{THCM-Cal}           & 0.8973   & 0.0820  & 12.1923  \\
-- w/o Proposition             & 0.8976   & 0.0975  & 10.2582  \\
-- w/o ICD            & 0.8991   & 0.0889  & 11.2478  \\
\bottomrule
\end{tabular}%
}
\caption{Conformal Prediction Metrics on MIMIC-III and MIMIC-IV (Cov.\,$\uparrow$: higher is better; MIW\,$\downarrow$: lower is better; IE\,$\uparrow$: higher is better)}
\label{tab:conformal_metrics}
\end{table}

\subsection{Ablation Study}
To quantify the contribution of each component, we perform an ablation study (Table~\ref{tab:main_exp}) by removing: (i) contextual BERT embeddings in favor of simple indexing (``w/o BERT''), (ii) the ICD‐side causal graph (``w/o ICD''), (iii) the proposition‐side causal graph (``w/o Proposition''), and (iv) the conformal calibration module (``w/o ConCalib''). In every case, omitting a component degrades performance across both datasets.
First, replacing BERT with one‐hot indexing (``w/o BERT'') yields the largest drop: on MIMIC‐III, Precision@20 falls from 21.47\% to 14.34\%, and on MIMIC‐IV from 18.66\% to 13.98\%. This underscores the necessity of rich contextual text representations for capturing fine‐grained clinical nuances.
Second, removing the ICD‐side causal graph (``w/o ICD'') reduces Recall@20 on MIMIC‐III from 33.16\% to 25.82\% and on MIMIC‐IV from 46.04\% to 36.58\%, indicating that explicit modeling of inter‐code dependencies is critical to retrieving less frequent but clinically important comorbidities.
Third, ablating the proposition‐side graph (``w/o Proposition'') lowers Precision@10 from 30.02\% to 26.07\% on MIMIC‐III and Recall@10 from 37.03\% to 28.99\% on MIMIC‐IV, demonstrating the value of narrative‐to‐code triggers for prioritizing key diagnoses.
Finally, skipping conformal calibration (``w/o ConCalib'') causes AUROC to drop modestly from 92.07\% to 91.84\% on MIMIC‐III and from 94.91\% to 94.33\% on MIMIC‐IV, confirming that score recalibration is important for maintaining stable discrimination under dataset shift.

% \subsection{Time Efficiency Comparasion}

\begin{figure}[!t] 
  \centering
  \includegraphics[width=0.38\textwidth]{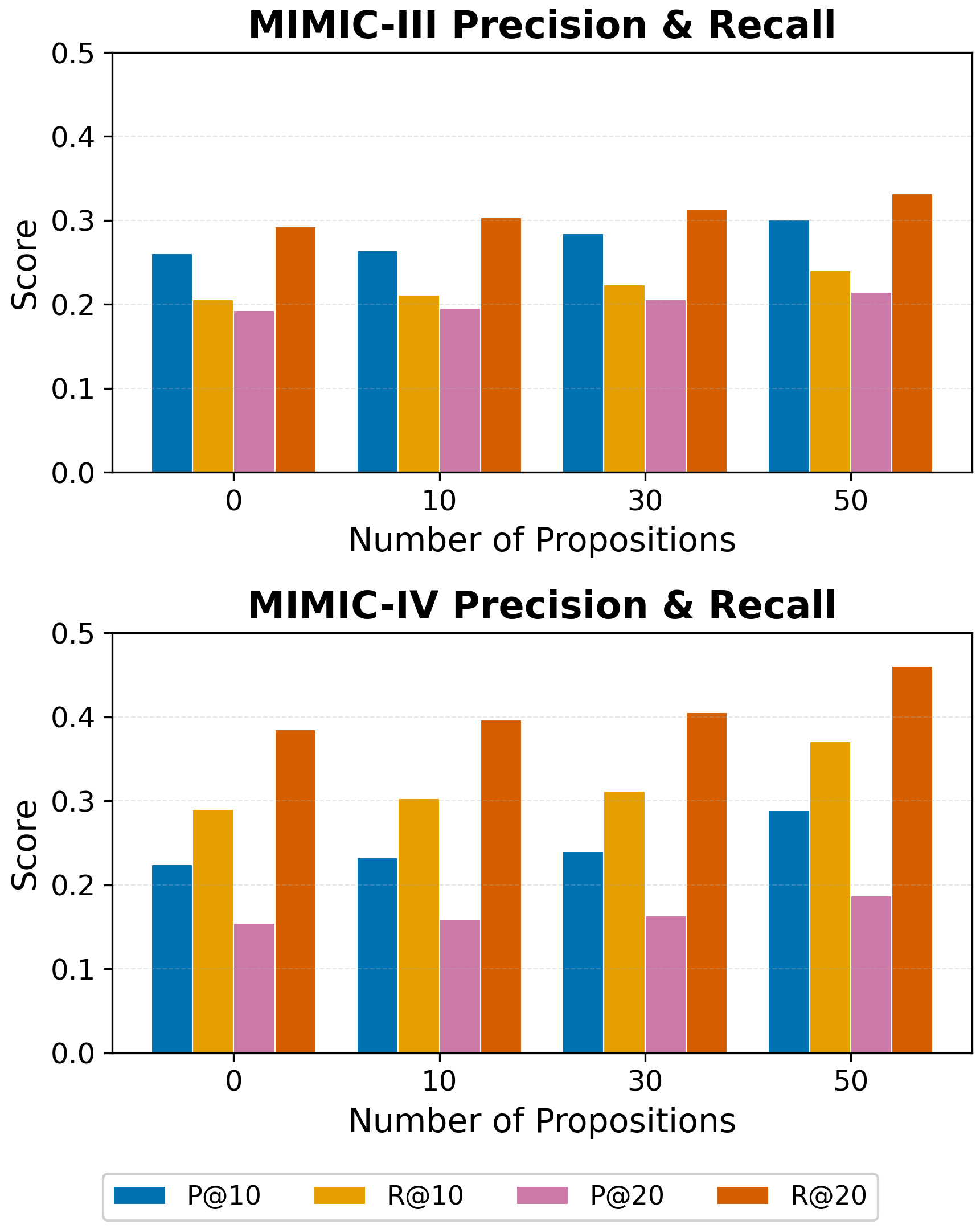}
  % \caption{Proposition Num Influence}
  \caption{Effect of the number of extracted propositions for multi‐label ICD‐9 prediction.}
  \label{fig:prop_num}
\end{figure}

\subsection{Analysis on Conformal Prediction}

We evaluate conformal prediction using three complementary metrics. Coverage (Cov) measures the fraction of true ICD codes captured by the prediction set. Mean interval width (MIW) quantifies the average size of these sets, with smaller widths indicating tighter intervals. Interval efficiency (IE), defined as the reciprocal of MIW, directly reflects interval compactness. Table~\ref{tab:conformal_metrics} presents these metrics, averaged over all epochs, for our full \textsc{THCM-Cal} model and two ablations on both MIMIC‐III and MIMIC‐IV.

As shown in Table~\ref{tab:conformal_metrics}, \textsc{THCM-Cal} attains the highest coverage and the narrowest intervals on both datasets, leading to superior efficiency. On MIMIC‐III the full model achieves a coverage of 0.9006 alongside a mean interval width of 0.1087 (IE = 9.1990), whereas removing the proposition side reduces coverage to 0.8980 and widens intervals to an average width of 0.1137, and omitting the ICD side further lowers coverage to 0.8937 with intervals of width 0.1144. Comparable patterns emerge on MIMIC‐IV, where \textsc{THCM-Cal} delivers efficiency of 12.1923 compared to 10.2582 without the proposition side and 11.2478 without the ICD side. These results demonstrate that the proposition side is crucial for concentrating uncertainty into tighter sets and that the ICD side preserves nominal coverage under label sparsity. Neither ablation matches the performance of the complete model, confirming that both sides contribute to the final calibration quality.

\subsection{Effect of Extracted Propositions}

Figure~\ref{fig:prop_num} examines how varying the number of extracted propositions per admission affects prediction performance. As the proposition count increases from 0 to 50, both precision and recall at \(K=10\) and \(K=20\) steadily improve on MIMIC-III and MIMIC-IV. On MIMIC-III, Precision@10 rises from 0.26 to 0.30 and Recall@20 from 0.29 to 0.33; on MIMIC-IV, Precision@10 climbs from 0.22 to 0.29 and Recall@20 from 0.38 to 0.46.  
These gains reflect that each additional proposition injects new, fine-grained clinical observations into our causal graph, enabling more accurate identification of both the highest‐priority codes (boosting Precision@10) and the broader set of relevant comorbidities (boosting Recall@20). 
With more narrative nodes, the inter-slice propagation and cross-modality trigger edges become better grounded, leading to smoother per‐admission embeddings and reduced variance in performance across patients. 
Notably, the marginal benefit from adding propositions begins to plateau beyond 30 propositions, suggesting a point of diminishing returns where most salient information has already been captured.  In practical, setting \(K=30\) propositions strikes a favorable balance between extraction cost and predictive performance.

\subsection{Effect of Calibration}

% \begin{figure}[!t] 
%   \centering
%   \includegraphics[width=0.4\textwidth]{latex/figures/AB-Ca1.png}
%   % \caption{Proposition Num Influence}
%   \caption{}
%   \label{fig:AB-Ca1}
% \end{figure}

% \begin{figure}[!t] 
%   \centering
%   \includegraphics[width=0.4\textwidth]{latex/figures/AB-Ca2.png}
%   % \caption{Proposition Num Influence}
%   \caption{}
%   \label{fig:AB-Ca2}
% \end{figure}

Figure~\ref{fig:AB-Ca-combined} compares models trained with versus without our split conformal calibration on MIMIC-IV. Across all four metrics and at every checkpoint (epochs 5, 10, 15, 20), the calibrated model achieves higher scores than the non-calibrated baseline, illustrating that calibration not only provides valid confidence intervals but also leads to more accurate top-K code retrieval. The larger gap at early epochs indicates that calibration rapidly corrects over- and under-confidence in raw probabilities, yielding stable and reliable predictions as training progresses. 
Beyond accuracy, conformal calibration also ensures that the empirical coverage of each code’s prediction set aligns with the desired level, even under non-stationary label co-occurrence patterns in MIMIC-IV. 
Importantly, calibration mitigates the impact of skewed code frequencies by adaptively adjusting thresholds per label, thereby reducing overprediction of common codes and underprediction of rare ones. These results confirm that split conformal calibration serves as a good post-hoc uncertainty quantifier.

% \subsection{Causalty Visialization}

\section{Conlcusion}

In this work, we introduced \textsc{THCM-Cal}, a unified Temporal–Hierarchical Causal Model with Conformal Calibration for Clinical Risk Prediction. By constructing a multi‐slice causal graph that jointly captures intra‐visit proposition sequencing, intra‐visit cross‐modality triggers, and inter‐visit risk propagation, our model uncovers clinically meaningful relationships between narrative observations and diagnostic codes. We further adapt split conformal prediction to the multi‐label setting, providing finite‐sample guarantees on per‐code. Extensive experiments on MIMIC‐III and MIMIC‐IV benchmarks demonstrate that \textsc{THCM-Cal} substantially outperforms state‐of‐the‐art baselines, while offering calibrated uncertainty estimates. Ablation studies confirm that each causal edge type and the conformal calibration module are critical to performance gains.

% Ablation studies confirm that each causal edge type and the conformal calibration module are critical to performance gains. Our results highlight the promise of integrating causal discovery and conformal inference for reliable, interpretable clinical decision support.  

% In future work, we plan to scale causal graph learning to longer admissions, incorporate expert‐in‐the‐loop interventions for edge validation, and extend our framework to ICD‐10 and multilingual clinical corpora. 

\section*{Acknowledgement}
We would like to thank the anonymous reviewers for their insightful comments and suggestions. This work was supported by the British Heart Foundation Manchester Research Excellence Award (RE/24/130017). We also acknowledge the CSF3 at the University of Manchester for providing GPU resources. Xin is supported by the UoM-CSC Joint Scholarship.

\clearpage

\section*{Limitations}

While \textsc{THCM-Cal} demonstrates strong gains in accuracy, interpretability, and uncertainty calibration, it has several limitations. 
\begin{itemize}
    \item First, the reliance on large language models (e.g., GPT-3.5) for atomic proposition extraction introduces additional computational overhead and may propagate errors when the extractor misidentifies or omits clinically relevant statements. 
    \item Second, although we validate on two MIMIC datasets mapped to ICD-9, extending \textsc{THCM-Cal} to richer coding systems (e.g., ICD-10) or to non‐English clinical corpora will require careful adaptation of both the proposition extraction and code description modules. Addressing these challenges is our future work.  
\end{itemize}
Despite these considerations, our approach provides a solid foundation for predicting ICD, and we believe that further refinements in these areas can further enhance its applicability.

\bibliography{THCM-Cal}

\clearpage
\appendix
\section{Thresholding Strategies for F1 Score Calculation}
In our experiments we do not report F1 scores because each baseline employs a different method to convert model scores into binary labels. This variation makes direct comparison of F1 values problematic. Instead, we provide AUROC, P@10, R@10, P@20 and R@20 which do not depend on a fixed threshold and thus allow a fair comparison. For reference we summarize below the thresholding strategies used in recent ICD code prediction works.

Specifically, \textbf{Chet}~\cite{lu2022context} counts the number of true labels for each instance and selects that many codes with the highest predicted scores. \textbf{CARER}~\cite{Nguyen2024CARER} for multi‐label tasks uses a dynamic threshold equal to the number of true labels and for binary tasks applies a fixed cutoff of 0.5. \textbf{DKEC}~\cite{Chen2023DKEC} applies a fixed cutoff of 0.5 and marks all codes with probability at least 0.5 as positive and if none meets this threshold it selects the single code with the highest probability. \textbf{CACHE}~\cite{Xu2022CACHE} uses the same 0.5 cutoff but allows this value to be adjusted at runtime. \textbf{RAM‐EHR}~\cite{DBLP:conf/acl/XuSYZJWHY24} applies a 0.5 threshold independently for each code and reports per‐label metrics. \textbf{HiTANet}~\cite{Luo2020HiTANet} does not use any threshold and instead selects exactly one code by choosing the label with the maximum model score.

As shown above some models use dynamic thresholds matching each sample’s true label count some use a fixed threshold of 0.5, and one uses maximum-probability selection. This heterogeneity in thresholding makes F1 scores difficult to compare on equal terms. We therefore omit F1 from our evaluations and rely on threshold-independent metrics to ensure a fair assessment of all methods.

\end{document}